\documentclass{article}
\usepackage{ijcai26}
\usepackage{verbatim}
\usepackage{amsfonts}
\usepackage{amssymb}
\usepackage{times}
\usepackage{soul}
\usepackage{url}
\usepackage[hidelinks]{hyperref}
\usepackage[utf8]{inputenc}
\usepackage[small]{caption}
\usepackage{graphicx}
\usepackage{amsmath}
\usepackage{amsthm}
\usepackage{booktabs}
\usepackage{algorithm}
\usepackage{algorithmic}
\usepackage{multirow}
\usepackage[switch]{lineno}

\title{TCDA: Thread-Constrained Discourse-Aware Modeling for Conversational Sentiment Quadruple Analysis}

\author{
Xinran Li$^1$
\and
Xinze Che$^1$\and
Yifan Lyu$^1$\and
Zhiqi Huang$^2$\And
Xiujuan Xu$^{1,}$\thanks{Corresponding author.}\\
\affiliations
$^1$School of Software, Dalian University of Technology, Dalian, China\\
$^2$School of Data and Computer Science, Sun Yat-sen University, Guangzhou, China
\emails
963707605@mail.dlut.edu.cn \and xjxu@dlut.edu.cn
}

\begin{document}

\maketitle

\begin{abstract}
    Conversational Aspect-based Sentiment Quadruple Analysis (DiaASQ) needs to capture the complex interrelationships in multiple rounds of dialogues. Existing methods usually employ simple Graph Convolutional Networks (GCN), which introduce structural noise and fail to consider the temporal sequence of the dialogues, or use standard RoPE, which implicitly captures relative distances in a flat sequence but cannot clearly separate the token-level syntactic order from the utterance-level progression, and may suffer from the Distance Dilution problem. To address these issues, we propose a new framework that combines Thread-Constrained Directed Acyclic Graph (TC-DAG) and Discourse-Aware Rotary Position Embedding (D-RoPE). Specifically, TC-DAG filters out cross-thread noise based on thread constraints, maintains global connectivity through root anchoring, and incorporates the temporal sequence of the dialogues. D-RoPE aligns multi-layer semantics using dual-stream projection and multi-scale frequency signals, captures thread dependencies using tree-like distances, and alleviates the token-level Distance Dilution problem by incorporating utterance-level progressions. Experimental results on two benchmark datasets demonstrate that our framework achieves state-of-the-art performance.
\end{abstract}

\section{Introduction}
\begin{figure}[t!]
\centering
\includegraphics[width=0.49\textwidth]{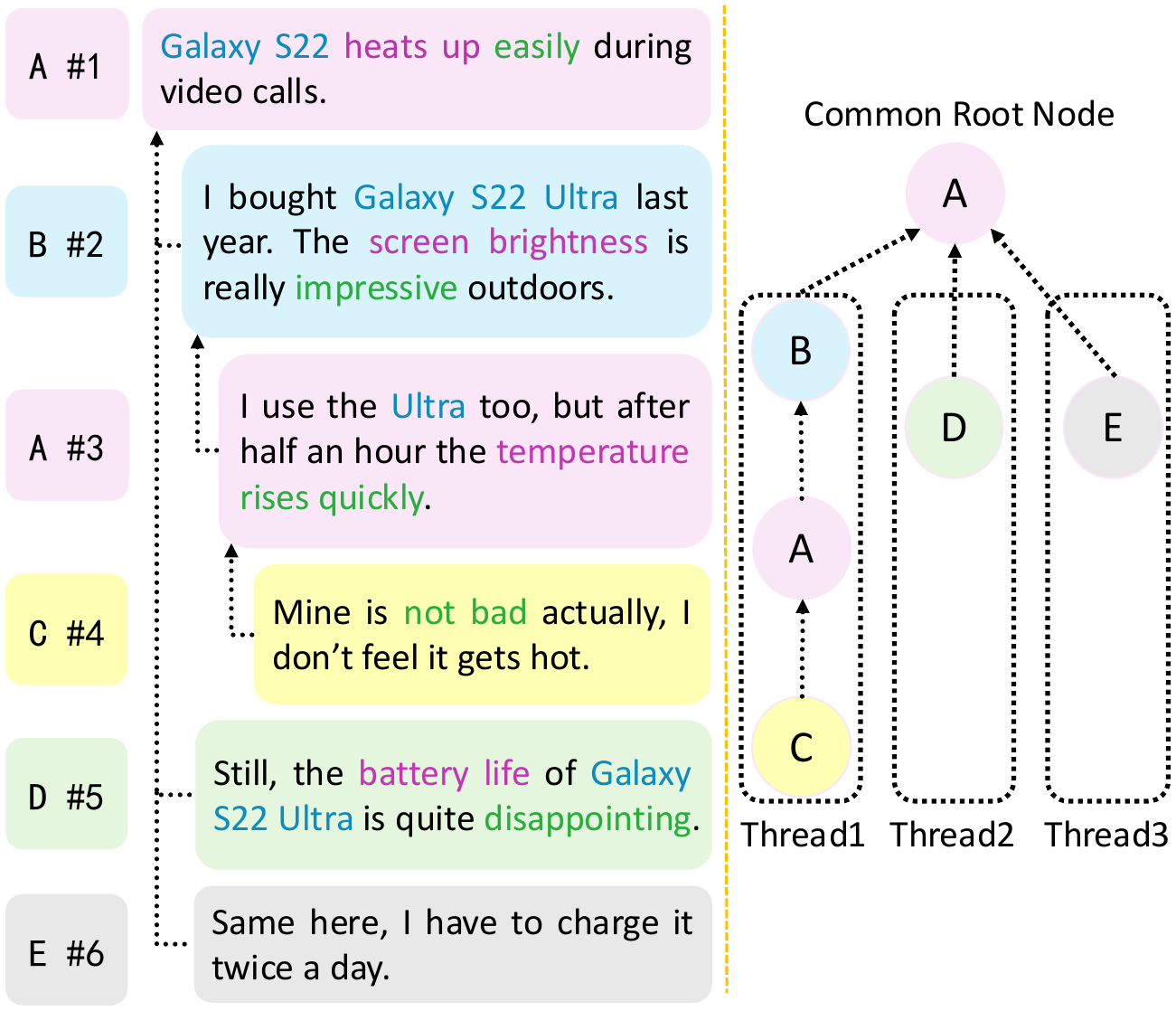} 
\caption{A sample dialogue and its corresponding thread structure.}
\label{fig1}
\end{figure}

With the rapid proliferation of online social media and real-time communication platforms, the task of Conversational Aspect-based Sentiment Quadruple Analysis (\textit{DiaASQ}) \cite{DiaASQ} has emerged to meet the growing demand for fine-grained sentiment understanding in conversations. As shown in Figure \ref{fig1} and Table \ref{tab:my_image_table}, the goal of \textit{DiaASQ} is to automatically extract all existing sentiment quadruples $(t, a, o, s)$ from the given multi-round conversation. In this formulation, \textbf{target $t$} (the object of discussion), \textbf{aspect $a$} (the specific attribute of the target) and \textbf{opinion $o$} (the subjective expression about the aspect) correspond to specific text spans in the conversation. Meanwhile, \textbf{sentiment $s$} represents the emotional polarity, which is usually classified as positive, negative or neutral. Different from traditional sentence-level sentiment analysis \cite{zhang-etal-2021-aspect-sentiment,mao-etal-2022-seq2path}, \textit{DiaASQ} faces significant challenges due to the fragmented nature of the information and the inherent complex context dependencies in the conversation context.

\begin{table}[t!]
\centering
\includegraphics[width=0.49\textwidth]{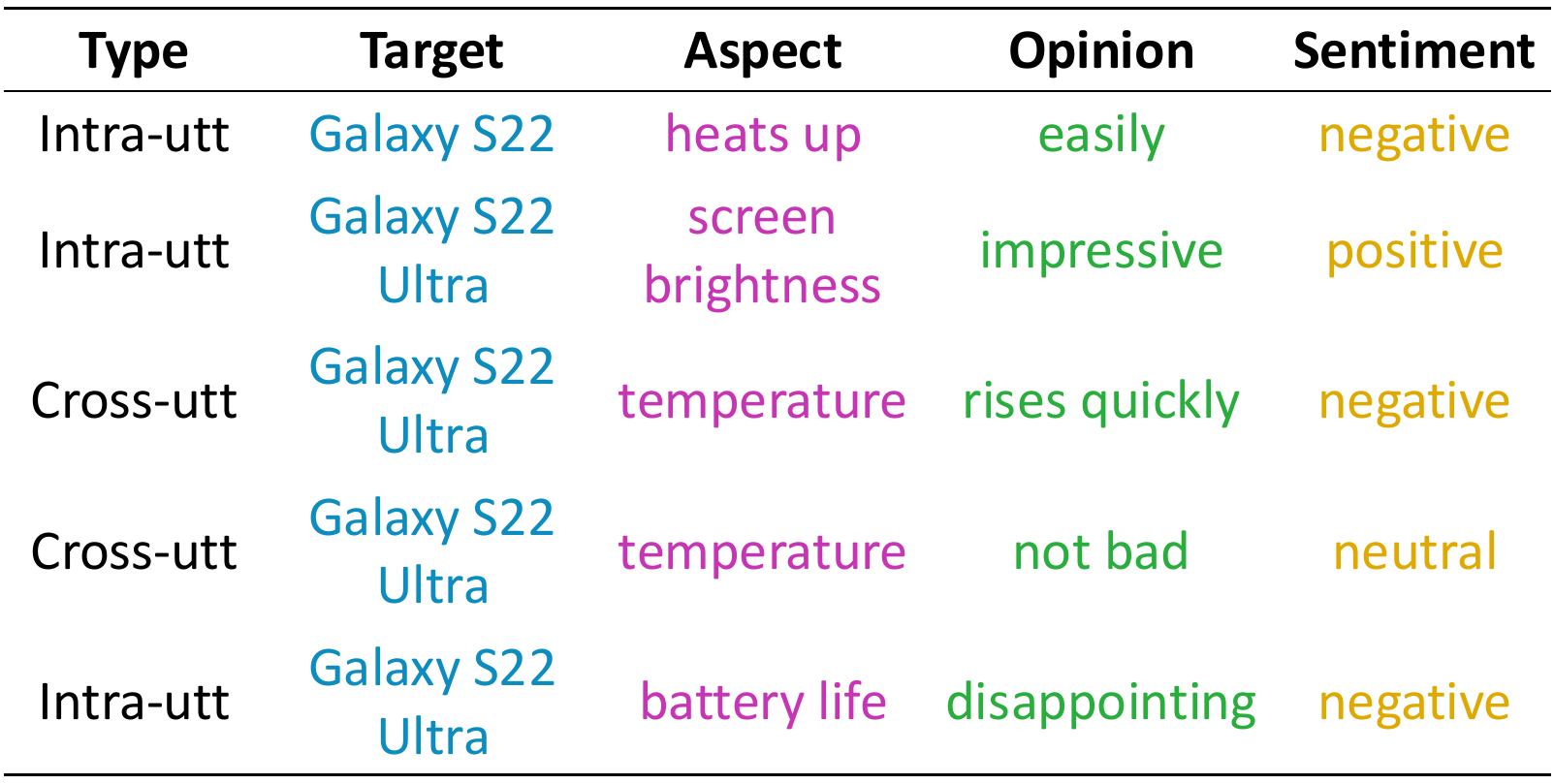} 
\caption{The sentiment quadruple annotations for the dialogue.}
\label{tab:my_image_table}
\end{table}

To capture the structural details of the conversation, DMIN \cite{DMIN} introduced the concept of “discourse thread structure". As shown in Figure \ref{fig1}, the conversation has a highly structured feature, consisting of multiple utterances and their corresponding speakers. These utterances can be decomposed into different semantic threads \cite{DMCA,thread1}. Under this framework, except for the root node, each utterance is closely related to a specific response target, forming a tree-like topological dependency relationship. This complex interaction pattern implies that the flow of sentiment is not only limited by the sequential arrangement of words but also by the topological structure of the conversation. Although the introduction of the thread structure has brought about performance improvements, the existing methods still have difficulty fully leveraging these complex dependencies. Specifically, current models \cite{H2DT,MARN,DMIN} typically use a general Graph Neural Networks (GCN) to handle the conversation structure, treating the reply-to relations as a simple edge \cite{RGCN,veličković_casanova_li&ograve;_cucurull_romero_bengio_2018}. However, current paradigms usually have two limitations. Firstly, they ignore the semantic isolation between independent threads, inevitably introducing structural noise from irrelevant threads. Secondly, they treat the dynamic conversation as a static graph structure, ignoring the natural temporal order and different speaker identities in the utterances. This failure to implement sequential and speaker-sensitive constraints results in the complex interaction between local context and overall discourse logic not being fully explored \cite{LSDGNN}.

To capture the relative distances between sentiment elements, Rotary Position Embedding (RoPE) \cite{RoPE} has been widely adopted in recent \textit{DiaASQ} frameworks \cite{DiaASQ,DMIN,H2DT}. However, existing implementations typically employ a fragmented and cumulative strategy, often restricting entity extraction to the local token context, or simply adding separate attention scores from the token and utterance levels. This token-based modeling introduces a key issue, which we call \textit{Distance Dilution}: in multi-round conversations, verbose utterances expand the distance between logically adjacent turns (e.g., a Q\&A pair separated by 50+ tokens). Under high-frequency RoPE rotations, this expanded distance causes the positional correlation to decay prematurely, cutting off semantic connections. Therefore, these mechanisms are difficult to balance both the high sensitivity to local syntax and the long-term retention ability for the global discourse simultaneously.

To address these challenges, we propose the TCDA framework, which integrates explicit topological structure and implicit positioning. Firstly, we introduce the Thread Constraint Directed Acyclic Graph (TC-DAG) to construct an accurate dialogue structure model. Unlike general GCNs that indiscriminately propagate information, TC-DAG sets strict thread-level boundaries. This design effectively suppresses structural noise from irrelevant branches while retaining the logical evolution from the root node to the leaf nodes. Secondly, we propose Discourse-Aware Rotary Position Embedding (D-RoPE) to alleviate the Distance Dilution and overcome the limitations of additive modeling. Unlike the standard encoding method that loosely couples local and global features through linear superposition, D-RoPE constructs a joint semantic-structural embedding. It projects tokens and utterances to independent subspaces and applies topology-adaptive coordinate transformation. This mechanism ensures that fine lexical cues and coarse discourse logic can be deeply integrated before the interaction, enabling accurate interpretation of cross-turn dependencies, regardless of intervening verbosity.

Our contributions can be summarized as follows:
\begin{itemize}
    \item We propose the Thread Constraint Directed Acyclic Graph (TC-DAG), which employs intra-thread constraints and a fixed root mechanism to suppress structural noise while preserving logical coherence.
    
    \item We introduce Discourse-Aware Rotary Position Embedding (D-RoPE), featuring dual-stream projections that decouple micro- and macro-semantics to mitigate distance dilution and align multi-scale distances.
    
    \item TCDA achieves SOTA performance. Our code is available at \url{https://github.com/LiXinran6/TCDA}.
\end{itemize}

\section{Related Work}
\subsection{Aspect-Based Sentiment Analysis}
Early studies on ABSA mainly focused on simple, isolated sentences with a single structure. Initially, they concentrated on single-element tasks such as aspect extraction \cite{AspectTE} and polarity classification \cite{dual-graph}. To obtain more comprehensive sentiment information, subsequent research shifted to compound tasks, including Aspect-Opinion Pair (AOPE) \cite{Wu2021LearnFS} and Triplet Extraction (ASTE) \cite{chen-etal-2022-enhanced,zhao-etal-2024-dual}, which aim to jointly identify aspect terms, opinion terms, and their corresponding polarities. Recently, to provide a comprehensive sentiment picture, the research focus has shifted to Aspect Sentiment Quadruple Prediction (ASQP) \cite{zhang-etal-2021-aspect-sentiment}. This task extracts the complete $(a, c, o, s)$ quadruple using predefined aspect categories $c$.

\begin{figure*}[t]
\centering
\includegraphics[width=0.95\textwidth]{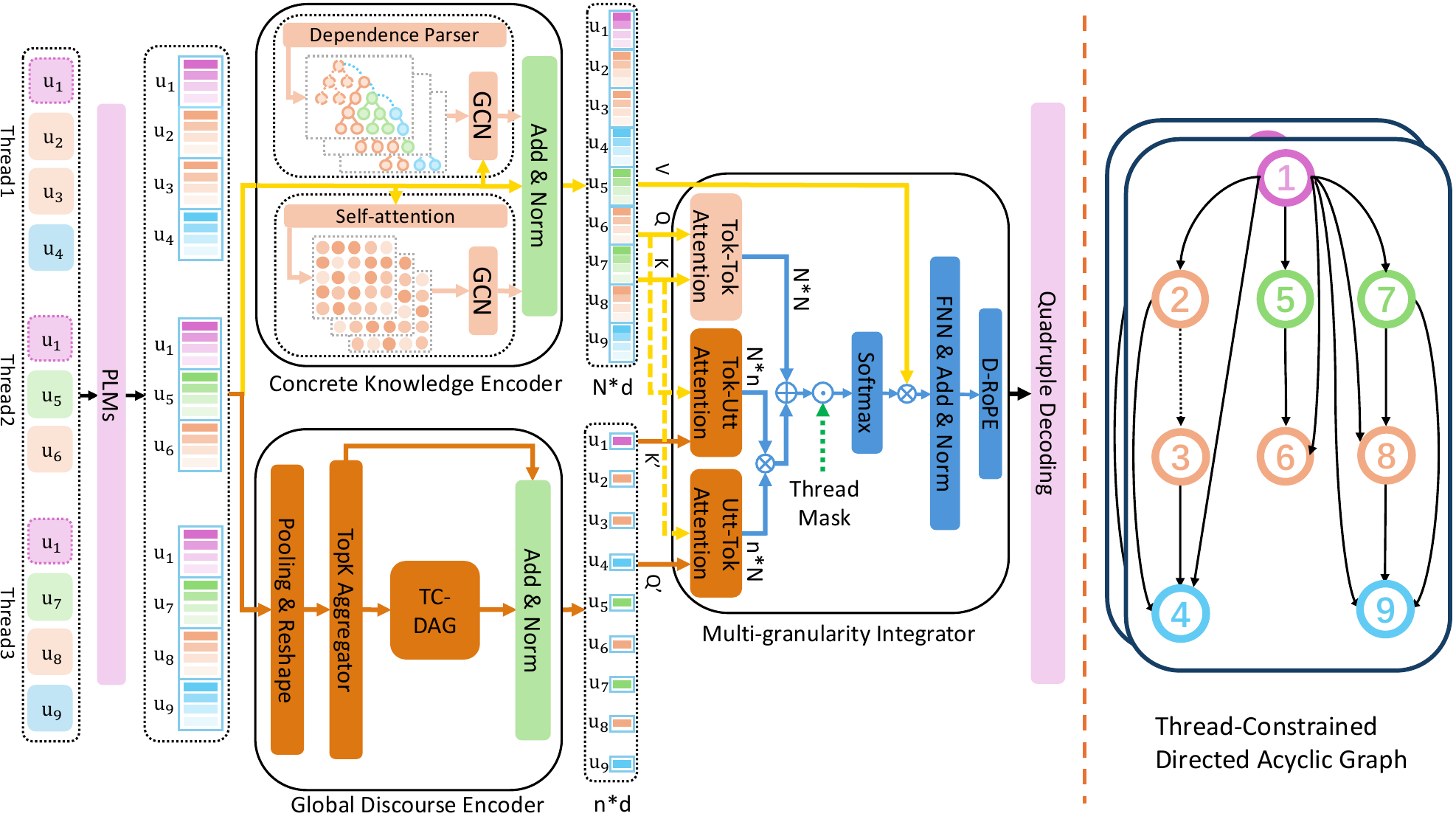} 
\caption{The overall architecture of our proposed TCDA.}
\label{structure}
\end{figure*}

\subsection{Conversational Aspect-Based Sentiment
Quadruple Analysis}

Although the traditional ABSA benchmarks mainly focus on sentence-level \cite{SemEval-2014,SemEval-2016}, they limit the applicability of existing methods in multi-turn conversation scenarios \cite{9996141}. To bridge this gap, the \textit{DiaASQ} task was introduced \cite{DiaASQ}, which employs three parallel attention matrices to explicitly capture the complex inter-utterance correlations. Subsequently, numerous studies further explored this task from different structural perspectives.

H2DT \cite{H2DT} employs a heterogeneous attention network and a ternary scorer to enhance the cohesion of quadruples, while DMCA \cite{DMCA} and ICMSR \cite{ICMSR} both utilize a multi-scale mechanism - specifically, windows and the SMM module - to capture long-range dependencies and structural features. Specifically, DMIN \cite{DMIN} is the first to use GCN and multi-granularity integration to incorporate thread structure, enabling token interactions to match the utterance-level discourse. Although CA-DAGNet \cite{CA-DAGNet} constructs a Directed Acyclic Graph \cite{DAGNN,DAG-ERC} to capture cross-utterance dependencies, it ignores the inherent thread-based topological constraints. Additionally, recent frameworks \cite{DiaASQ,H2DT} have integrated RoPE \cite{RoPE} to encode relative distances within the conversational tree. However, these RoPE implementations are typically limited to encoding the local token context or adopting a fragmented strategy of simple linear superposition, which ignores the differences in frequency scales and cannot alleviate the Distance Dilution caused by verbose utterances.

\section{Methodology}
We propose TCDA, which combines TC-DAG and D-RoPE. Its overall architecture is shown in Figure \ref{structure}.

\subsection{Problem Definition}
In the \textit{DiaASQ} task, each conversation is represented as $D = \{u_1, u_2, \dots, u_n\}$, along with the reply index set $R = \{l_1, l_2, \dots, l_n\}$ and the speaker sequence $S = \{s_1, s_2, \dots, s_n\}$. Here, $l_i$ indicates that the utterance $u_i$ is a direct response to $u_{l_i}$. Each utterance $u_i = \{w_{1}, \dots, w_{m_i}\}$ consists of $m_i$ tokens.

Following the grid tagging framework \cite{DiaASQ,DMIN}, we rephrase the extraction of the quadruple as a unified relation tagging problem. For any pair of words $(w_a, w_b)$ in the flattened dialogue, the model is trained to identify three types of semantic connections:

\begin{itemize}
\item \textbf{Entity Boundaries} ($y_{ent} \in \{\text{TGT, ASP, OPI}\}$): These labels define the corresponding range by connecting the start and end tokens of the target, aspect, and opinion. For example, the TGT association from “iPhone" to “14" will identify “iPhone 14" as a target entity.

\item \textbf{Entity Alignment} ($y_{pair} \in \{\text{H2H, T2T}\}$): These relationships link different entities together. Specifically, \textit{head-to-head} (H2H) and \textit{tail-to-tail} (T2T) tags are used to pair the entities, for example, associating the target “iPhone 14" with its corresponding aspect “battery life".

\item \textbf{Sentiment Polarity} ($y_{pol} \in \{\text{POS, NEG, NEU}\}$): This value indicates the sentiment tendency (positive, negative, or neutral) between the related entities.

\end{itemize}
For each sub-task, if there is no specific relationship between these tokens, a special label \textit{other} will be assigned to it.

\subsection{Textual Feature Extraction}
Inspired by DMIN \cite{DMIN}, each conversation is divided into multiple threads $T_k = \{u'_1, u'_i,  u'_{i+1}, \dots, u'_j\}$, starting from a common root node $u'_1$, to balance the context window limit on PLM and the discourse interaction. As shown in Figure \ref{fig1}, threads are arranged in sequence and only cross at the root node. Each utterance is formatted as $u'_i = \{[\text{CLS}], u_i, s_i\}$ to incorporate speaker information. The encoding form at the thread level is: 
\begin{equation}
H_{T_k} = \{H_1^{u'}, H_i^{u'}, \dots, H_j^{u'}\} = \text{PLM}(T_k),
\end{equation}
where $H_i^{u'} = \{h_{i}^{\text{cls}}, H_i^{u}, h_{i}^{s}\}$ contains token features $H_i^{u} \in \mathbb{R}^{m_i \times d}$.

\subsection{Dual-scale Contextual Encoding}
To simultaneously capture fine-grained semantic cues and coarse-grained discourse structure, we propose a dual-scale encoding framework. This module refines the text representation by performing knowledge enhancement at the thread level and discourse modeling at the conversation level.

\paragraph{Token-level Knowledge Encoding. }
In order to strike a balance between global and local interactions within the PLM context window, we first perform knowledge enhancement within each individual thread $T_k$. Following \cite{DMIN}, we employ a structure called Concrete Knowledge Encoder (CKEncoder), which consists of parallel Syntactic and Semantic GCNs \cite{GCN1,GCN2,Syntactic,Sem}. Specifically, we extract local knowledge features $\tilde{H}_{T_k}$ based solely on the thread-specific context to filter out cross-thread noise:
\begin{equation}
\tilde{H}_{T_k} = \sum_{g \in \{syn, sem\}} \text{GCN}_g(A_g, H_{T_k}),
\end{equation}
where $A_{syn}$ and $A_{sem}$ respectively represent the thread-level syntactic and semantic adjacency matrices. Subsequently, we aggregate the original features $H_{T_k}$ and the knowledge features $\tilde{H}_{T_k}$ from all threads to reconstruct their global corresponding features $H_{tok}$ and $\tilde{H}_{tok}$ (by averaging the shared root node $u'_1$). The final enhanced token representation $H'_{tok}$ is obtained through global residual connections and layer normalization: 
\begin{equation}
H'_{tok} = \text{LN}(H_{tok} + \tilde{H}_{tok}).
\end{equation}

\paragraph{Utterance-level Discourse Modeling.} 
Meanwhile, we abstract the original global token-level feature $H_{tok}$ into an utterance-level representation $H_{utt} = \{h_1, \dots, h_n\}$ through a Top-K aggregator \cite{DMIN}. These representations can capture the flow of the conversation, but require powerful structural modeling. Unlike the previous methods that used fully connected graphs, we process $H_{utt}$ using a Thread-Constrained DAG (TC-DAG) to strictly follow the temporal order and replying topology of the conversation. For more details, please refer to Section \ref{DAG}.

\subsection{Thread-Constrained DAG} 
\label{DAG}
To strictly adhere to the dialogue structure and filter out irrelevant information, we propose the Thread-Constrained Directed Acyclic Graph (TC-DAG), which is represented as $\mathcal{G} = (\mathcal{V}, \mathcal{E}, \mathcal{R})$. Here, $\mathcal{V}$ represents the utterances, and there is a directed edge $(u_j \to u_i)$ only when $j < i$. The relation set $\mathcal{R} = \{0, 1\}$ indicates whether the connected nodes were uttered by the same speaker.

\begin{figure}[t]
\centering
\includegraphics[width=0.48\textwidth]{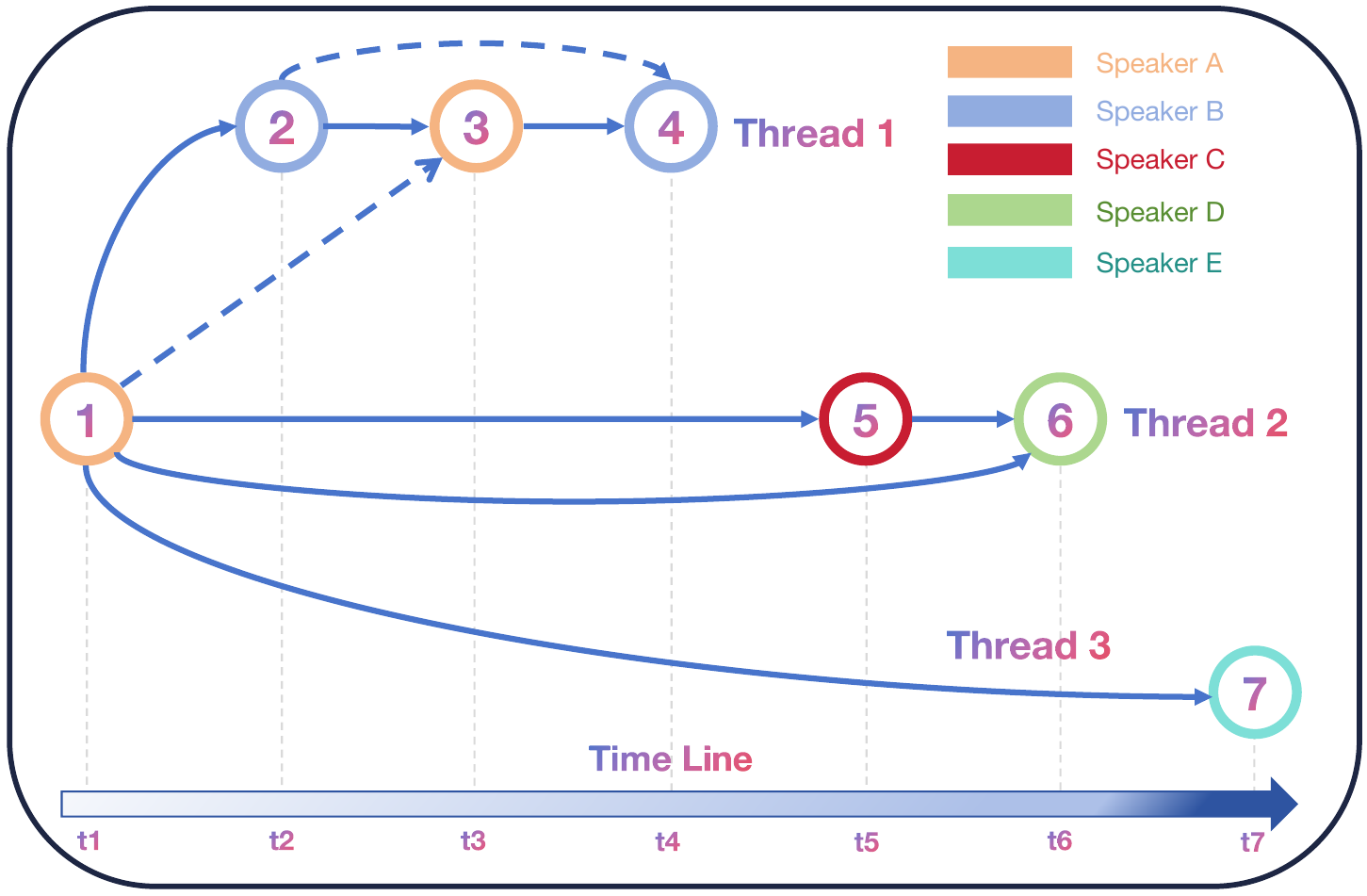} 
\caption{TC-DAG construction ($\omega=1$). Solid/dashed arrows denote inter- and same-speaker dependencies among chronological utterances. The structure incorporates Global Root Accessibility, allowing nodes in divergent threads (e.g., $u_6, u_7$) to connect to the global root $u_1$ under the window constraint.}
\label{fig:dag_construction}
\end{figure}

\begin{algorithm}[t]
\caption{Building a Thread-Constrained DAG}
\label{alg:tc_dag}
\begin{algorithmic}[1]
\REQUIRE Dialogue $\{u_1, \dots, u_N\}$, speaker $P(\cdot)$, thread mapping $T(\cdot)$, window size $\omega$.
\ENSURE Graph $\mathcal{G} = (\mathcal{V}, \mathcal{E}, \mathcal{R})$.
\STATE $\mathcal{V} \leftarrow \{u_1, \dots, u_N\}, \mathcal{E} \leftarrow \emptyset, \mathcal{R} \leftarrow \{0, 1\}$
\FOR{$i = 2$ \TO $N$}
    \STATE $c \leftarrow 0, \quad \tau \leftarrow i - 1$
    \STATE $S_i \leftarrow \text{Start index of thread } T(u_i)$
    \WHILE{$\tau \ge S_i$ \AND $c < \omega$}
        \STATE $r \leftarrow (P(u_\tau) = P(u_i)) ? 1 : 0$
        \STATE $\mathcal{E} \leftarrow \mathcal{E} \cup \{(u_\tau, u_i, r)\}$
        \IF{$r = 1$} \STATE $c \leftarrow c + 1$ \ENDIF
        \STATE $\tau \leftarrow \tau - 1$
    \ENDWHILE
    \IF{$c < \omega$ \AND $S_i > 1$}
        \STATE $r \leftarrow (P(u_1) = P(u_i)) ? 1 : 0$
        \STATE $\mathcal{E} \leftarrow \mathcal{E} \cup \{(u_1, u_i, r)\}$
    \ENDIF
\ENDFOR
\RETURN $\mathcal{G} = (\mathcal{V}, \mathcal{E}, \mathcal{R})$
\end{algorithmic}
\end{algorithm}

\subsubsection{Constructing a Graph through Conversation}
A \textit{thread} refers to a sequence within a local conversation branch. To filter out structural noise, TC-DAG employs a retrospective strategy to limit the connection range of edges to be within these threads: each node is connected to the previous utterance that covers $\omega$ instances from the same speaker, including all intermediate background information. To ensure global connectivity, when reaching the thread boundary within the window, the connection extends to the root node $u_1$. This process organizes the conversation into a tree-like DAG (see Figure \ref{fig:dag_construction} and Algorithm \ref{alg:tc_dag}).

\subsubsection{Structure-Aware Relational Encoding}
Based on the constructed TC-DAG and the initial utterance feature $H_{utt}$, we use a relational GNN to propagate context information along the topological structure. Unlike the standard GNNs, which uniformly aggregates neighbors, our model specifically considers the sequential nature of the conversation and the different dependency types defined in $\mathcal{R}$. Let $\mathbf{h}_i^{(l)}$ represent the hidden state of utterance $u_i$ in the $l$-th layer, where the input state $\mathbf{h}_i^{(0)}$ corresponds to the vector $\mathbf{h}_i \in H_{utt}$. Since the DAG is strictly arranged in chronological order, we update the nodes from $i=1$ to $n$ sequentially. This ensures that when calculating $u_i$, the updated states $\mathbf{h}_j^{(l)}$ of all predecessor utterances $u_j \in \mathcal{N}_i$ (where $j < i$) are already available.

For a specific node $u_i$, the information aggregation is computed via a relation-aware attention mechanism. The attention coefficient $\alpha_{ij}^{(l)}$ for a neighbor $u_j \in \mathcal{N}_i$ is calculated as:
\begin{align}
    \alpha_{ij}^{(l)} &= \text{Softmax}_{j \in \mathcal{N}_i} \left( \mathbf{W}_\alpha^{(l)} \left[ \mathbf{h}_j^{(l)} \, \| \, \mathbf{h}_i^{(l-1)} \right] \right)
    \label{eq:attention}
\end{align}
where $\|$ denotes concatenation. The context vector $\mathbf{m}_i^{(l)}$ is then derived by:
\begin{align}
    \mathbf{m}_i^{(l)} &= \sum_{j \in \mathcal{N}_i} \alpha_{ij}^{(l)} \mathbf{W}_{r_{ij}}^{(l)} \mathbf{h}_j^{(l)}
    \label{eq:aggregation}
\end{align}
where $\mathbf{W}_{r_{ij}}^{(l)} \in \{\mathbf{W}_0^{(l)}, \mathbf{W}_1^{(l)}\}$ is a relation-specific projection matrix selected based on whether $u_i$ and $u_j$ share the same speaker ($r_{ij} \in \mathcal{R}$). This allows the model to differentially weigh intra-speaker and inter-speaker dependencies.

In order to effectively integrate the aggregated contextual information with the node's own historical records, we adopt a dual gated update mechanism \cite{DAG-ERC}. Specifically, we employ two parallel GRU units to capture complementary information flows. The \textit{node update unit} ($\text{GRU}_H$) uses the context as guidance to update the node's state, while the \textit{context update unit} ($\text{GRU}_C$) models the evolution of the context:
\begin{align}
    \tilde{\mathbf{h}}_i^{(l)} &= \text{GRU}_H(\mathbf{h}_i^{(l-1)}, \mathbf{m}_i^{(l)}) \label{eq:gru_h} \\
    \mathbf{c}_i^{(l)} &= \text{GRU}_C(\mathbf{m}_i^{(l)}, \mathbf{h}_i^{(l-1)}) \label{eq:gru_c}
\end{align}
Here, the inputs and hidden states are logically swapped between the two GRUs to maximize feature interaction. Finally, the updated representation for node $u_i$ at layer $l$ is obtained by summing the outputs:
\begin{align}
    \mathbf{h}_i^{(l)} &= \tilde{\mathbf{h}}_i^{(l)} + \mathbf{c}_i^{(l)}
    \label{eq:final_update}
\end{align}
Finally, we extract the node states $H_{utt}^{(L)} = \{\mathbf{h}_1^{(L)}, \dots, \mathbf{h}_n^{(L)}\}$ from the last layer $L$ and apply a residual connection followed by layer normalization to yield the final global representations:
\begin{equation}
    H'_{utt} = \text{LN}(H_{utt} + H_{utt}^{(L)}).
\end{equation}

\subsection{Global-Local Interaction and Discourse-Aware Position Encoding}
After obtaining the global structure-aware representation $H'_{utt}$ through the TC-DAG module, our aim is to reintegrate this global background information into the token-level features and enhance the position sensitivity.

\subsubsection{Global-Local Interaction}
To bridge the gap between the coarse-grained discourse structure and the fine-grained token features, we employ the cross-attention mechanism. Token representation $H'_{tok}$ is used as the \textit{query}, while the global utterance representation $H'_{utt}$ serves as the \textit{key} and \textit{value}, enabling tokens to focus on the relevant discourse context and generate the comprehensive representation $H_{final}$.

\subsubsection{Discourse-Aware Rotary Position Embedding (D-RoPE)}
\label{sec:D-RoPE}

To alleviate the inherent \textit{Distance Dilution} phenomenon in the RoPE strategy, our D-RoPE method explicitly separates the semantic granularity into independent subspaces and fuses them before interaction.

\begin{algorithm}[t]
\caption{D-RoPE-Enhanced Attention}
\label{alg:D-RoPE}
\begin{algorithmic}[1]
\REQUIRE Queries/Keys $\mathbf{Q}, \mathbf{K}$ for tok/utt; Indices $P_{tok}, P_{utt}$.
\ENSURE Score matrix $\mathbf{S}$.
\STATE \textbf{Projection:} $\mathbf{V}^{mic/mac} \leftarrow \text{Linear}_{mic/mac}(\mathbf{V}^{tok/utt})$ for $\mathbf{V} \in \{\mathbf{Q}, \mathbf{K}\}$
\STATE \textbf{Coordinate Transform:} For pair $(i,j)$, let $\sigma_{ij} = -1$ if threads diverge else $1$. Set $\hat{p}^{(j)} \leftarrow \sigma_{ij} P^{(j)}$.
\STATE \textbf{Dual-Scale Rotation:} 
\STATE \quad $\tilde{\mathbf{q}}_i \leftarrow \text{Concat}(\mathcal{R}(\mathbf{q}_i^{mic}, P_{tok}^{(i)}), \mathcal{R}(\mathbf{q}_i^{mac}, P_{utt}^{(i)}))$
\STATE \quad $\tilde{\mathbf{k}}_j \leftarrow \text{Concat}(\mathcal{R}(\mathbf{k}_j^{mic}, \hat{p}_{tok}^{(j)}), \mathcal{R}(\mathbf{k}_j^{mac}, \hat{p}_{utt}^{(j)}))$
\RETURN $\mathbf{S} \text{ where } \mathbf{S}_{ij} = \tilde{\mathbf{q}}_i^\top \tilde{\mathbf{k}}_j$
\end{algorithmic}
\end{algorithm}

\paragraph{Dual-Scale Semantic-Structural Projection.}
We decompose the integrated representation $H_{final}$ into parallel tokens ( $\mathbf{h}_{tok}$ ) and utterances ( $\mathbf{h}_{utt}$ ) streams, and then project them onto separate subspaces:
\begin{align}
    \mathbf{q}_i^{mic} &= \mathbf{W}_{mic} \mathbf{h}_{tok, i}, \quad \mathbf{q}_i^{mac} = \mathbf{W}_{mac} \mathbf{h}_{utt, i} \\
    \mathbf{k}_j^{mic} &= \mathbf{W}_{mic} \mathbf{h}_{tok, j}, \quad \mathbf{k}_j^{mac} = \mathbf{W}_{mac} \mathbf{h}_{utt, j}
\end{align}
where $\mathbf{W}_{mic}$ and $\mathbf{W}_{mac}$ are learnable matrices that separate local syntactic cues from the global discourse semantics.

\begin{table}[t]
\centering
\small
\setlength{\tabcolsep}{5pt} 
\begin{tabular}{ll cc ccc}
\toprule
\textbf{Data} & \textbf{Set} & \textbf{$D$} & \textbf{$U$} & \textbf{$Q_{tot}$} & \textbf{$Q_{int}$} & \textbf{$Q_{cro}$} \\
\midrule
\multirow{4}{*}{\textbf{ZH}} 
& Train & 800 & 5,947 & 4,607 & 3,594 & 1,013 \\
& Valid & 100 & 748   & 577   & 440   & 137 \\
& Test  & 100 & 757   & 558   & 433   & 125 \\
\cmidrule(lr){2-7}
& Total & 1,000 & 7,452 & 5,742 & 4,467 & 1,275 \\
\midrule
\multirow{4}{*}{\textbf{EN}} 
& Train & 800 & 5,947 & 4,414 & 3,442 & 972 \\
& Valid & 100 & 748   & 555   & 423   & 132 \\
& Test  & 100 & 757   & 545   & 422   & 123 \\
\cmidrule(lr){2-7}
& Total & 1,000 & 7,452 & 5,514 & 4,287 & 1,227 \\
\bottomrule
\end{tabular}
\caption{Statistics of ZH and EN datasets.}
\label{tab:dataset_stats}
\end{table}

\paragraph{Topology-Adaptive Rotary Encoding.}
We employ RoPE method with different base frequencies to encode the topological structure. While maintaining the standard relative position property $\tilde{\mathbf{q}}^\top \tilde{\mathbf{k}} = \mathbf{q}^\top \mathcal{R}(p_q - p_k) \mathbf{k}$, we introduce a Topology-Adaptive Coordinate Transformation that is applicable at both the micro and macro levels:

\begin{table*}[t]
\centering
\setlength{\tabcolsep}{12pt} 
\begin{tabular}{ll ccc cc}
\toprule
\multirow{2}{*}{\textbf{Data}} & \multirow{2}{*}{\textbf{Model}} & \multicolumn{3}{c}{\textbf{Pair Extraction (F1)}} & \multicolumn{2}{c}{\textbf{Quadruple (F1)}} \\
\cmidrule(lr){3-5} \cmidrule(lr){6-7}
& & \textbf{T-A} & \textbf{T-O} & \textbf{A-O} & \textbf{Micro} & \textbf{Ident.} \\
\midrule
\multirow{7}{*}{\textbf{ZH}} 
& MVQPN$^*$ \cite{DiaASQ} & 50.07	&50.40	&51.91 & 35.68	&41.37 \\
& H2DT$^*$ \cite{H2DT} & 50.00 & 48.20 & 52.56 & 39.85 & 43.03 \\
& DMCA \cite{DMCA} & 56.88  & 51.70  & 52.80  &  42.68 & 45.36 \\
& DMIN$^*$ \cite{DMIN} & \underbar{57.61} & 51.18  & \textbf{55.58}  &  \underbar{43.29} & \underbar{46.02} \\
& CA-DAGNet \cite{CA-DAGNet} & \textbf{58.76}  & 51.45   & 52.65   & 42.47   &45.44 \\
&IFusionQuad \cite{IFusionQuad} &54.68	&51.81	&50.04	&41.53	&44.56   \\
&ICMSR \cite{ICMSR}  &56.69 &\textbf{52.53} &54.59 &42.55 &45.20 \\
\cmidrule(lr){2-7}
&\textbf{TCDA (Ours)}  &57.47	&\underbar{52.24}	&\underbar{55.46}  &\textbf{44.35}	&\textbf{46.23}\\

\midrule
\multirow{7}{*}{\textbf{EN}} 
& MVQPN$^*$ \cite{DiaASQ} &48.60	&48.31 &50.05 & 35.62	&38.86 \\
& H2DT$^*$ \cite{H2DT} & 48.41   & 49.14  & 51.87  & 38.76  &  41.95 \\ 
& DMCA \cite{DMCA} & 53.08  & 50.99  & \textbf{52.40}  & 37.96  &  41.00 \\
& DMIN$^*$ \cite{DMIN}  &  53.72  &  52.40 &  52.22 & 38.14  &41.85\\
& CA-DAGNet \cite{CA-DAGNet} & 54.04  &51.88   &50.97   & 37.87  & 41.52  \\
&IFusionQuad \cite{IFusionQuad} &52.65	&51.82	&51.94	&35.96	&41.49   \\
&ICMSR \cite{ICMSR} &\underbar{54.23} &\underbar{52.67} &51.61 &\underbar{39.36} &\textbf{44.06} \\
\cmidrule(lr){2-7}
&\textbf{TCDA (Ours)} &\textbf{55.40}	&\textbf{52.90}	&\underbar{52.29}   &\textbf{39.69}	&\underbar{42.12} \\
\bottomrule
\end{tabular}
\caption{Performance on DiaASQ. T/A/O denote Target, Aspect, and Opinion; Ident. is Identification F1. Results with $^*$ are our reproductions; others are cited. Best and second-best results are bolded and underlined.}
\label{tab:main_results}
\end{table*}

\begin{enumerate}
    \item \textbf{Micro-RoPE (Token Level)}: With a standard frequency $\theta_{mic} = 10000$, we define the token index $p_{tok}$ as the cumulative topological distance starting from the global root node \cite{DiaASQ}. To make the subtraction mechanism of RoPE compatible with the addition distance (i.e., $p_{tok}^{(i)} + p_{tok}^{(j)}$ between different branch threads), we apply the coordinate sign inversion: 
    \begin{equation}
        \hat{p}_{tok}^{(j)} = 
        \begin{cases} 
        p_{tok}^{(j)} & \text{if } x_i, x_j \text{ in same thread} \\
        -p_{tok}^{(j)} & \text{if } x_i, x_j \text{ in divergent threads}
        \end{cases}
    \end{equation}
    This transformation enables $\mathcal{R}(p_{tok}^{(i)} - \hat{p}_{tok}^{(j)})$ to accurately encode the topological path lengths between different threads, while preserving the linear relative distances within the same thread.

    \item \textbf{Macro RoPE (Utterance Level)}:
    Relying solely on token indexing can lead to \textit{distance dilution}, where verbose utterances increase the distance and disrupt the semantic connections under high-frequency rotation. To alleviate this, we introduce Macro-RoPE, using utterance-level index $p_{utt}$, with the base frequency $\theta_{mac} = 100$ reduced. This transformation preserves strong attention on logical dependencies:
    \begin{equation}
        \hat{p}_{utt}^{(j)} = 
        \begin{cases} 
        p_{utt}^{(j)} & \text{if } x_i, x_j \text{ in same thread} \\
        -p_{utt}^{(j)} & \text{if } x_i, x_j \text{ in divergent threads}
    \end{cases}
    \end{equation}
    This ensures constant turn-level distances, serving as a robust discourse anchor.
\end{enumerate}

\paragraph{Fusion.}
We construct a unified feature vector by concatenating the rotation embeddings of the two subspaces: 
\begin{align}
    \tilde{\mathbf{q}}_i &= [\tilde{\mathbf{q}}_i^{mic} \, \| \, \tilde{\mathbf{q}}_i^{mac}] \\
    \tilde{\mathbf{k}}_j &= [\tilde{\mathbf{k}}_j^{mic} \, \| \, \tilde{\mathbf{k}}_j^{mac}]
\end{align}
Here, $[\cdot \, \| \, \cdot]$ represents concatenation. Then, the topological adaptive score is calculated through the dot product: \begin{equation}
    \text{Score}(x_i, x_j) = \tilde{\mathbf{q}}_i^\top \tilde{\mathbf{k}}_j
    \label{eq:D-RoPE_score}
\end{equation}
This ensures dual-scale semantic and positional consistency.

\subsection{Quadruple Decoding and Learning}
To isolate the semantic influence, we project the $H_{final}$ item into three task-specific spaces ($S_{ent}$, $S_{rel}$, $S_{pol}$). We apply D-RoPE to each grid $g$ to derive topology-adaptive probabilities by Softmax:
\begin{equation}
    P(y_{ij}^g | x_i, x_j) = \text{Softmax}(\text{Score}_g(x_i, x_j))
\end{equation}
We minimize weighted cross-entropy loss:
\begin{equation}
    \mathcal{L} = - \sum_{g} \sum_{i,j} \alpha_{ij}^g \log P(y_{ij}^g | x_i, x_j)
\end{equation}
where $y_{ij}^g$ represents the true label, while $\alpha_{ij}^g$ denotes the category weight.

\section{Experiments and Analysis}

\subsection{Dataset and Implementation Details}
\paragraph{Dataset.} We conduct experiments on the Chinese (ZH) and English (EN) datasets \cite{DiaASQ}. The detailed statistics are presented in Table~\ref{tab:dataset_stats}.

\paragraph{Implementation Details.} Following existing methods, we use RoBERTa-Large \cite{RoBERTaAR} and Chinese-RoBERTa-wwm-ext-base \cite{Chinese} as backbones for EN and ZH, with Top-$K$ ratios $\lambda$ of 0.5 and 0.8, respectively. Both syntactic and semantic GCNs consist of 3 layers, while the TC-DAG has 2 layers. We employ a sliding window of size $w=3$. We train with a batch size of 2 and a 0.1 dropout rate. The AdamW optimizer is used with learning rates of 1e-5 for PLMs and 1e-4 for other parameters. All experiments are conducted on a single NVIDIA GeForce RTX 4090 GPU. All results, including baseline comparisons and ablation studies, are reported as the average of five independent runs to ensure statistical significance.

\subsection{Baselines}
We compare TCDA against several state-of-the-art baselines: 
 MVQPN \cite{DiaASQ} (the pioneering grid-tagging baseline), H2DT \cite{H2DT}, DMCA \cite{DMCA}, DMIN \cite{DMIN}, CA-DAGNet \cite{CA-DAGNet}, IFusionQuad \cite{IFusionQuad} and ICMSR \cite{ICMSR}. 

\subsection{Main Results}
Table \ref{tab:main_results} shows that TCDA achieves SOTA or competitive performance across all benchmarks.

\subsection{Ablation Study}
To assess the contribution of each component, we compare TCDA with three variants:
(1) \textbf{w/o TC-DAG}, replacing the thread-constrained topology with the standard reply-based GCN;
(2) \textbf{w/o D-RoPE}, replacing the Discourse-Aware positioning with the standard RoPE;
(3) \textbf{w/o Both}, removing both modules.

Table \ref{tab:ablation} shows that removing any component degrades performance, with the sharpest decline when both are absent. This confirms that TC-DAG and D-RoPE provide complementary benefits in filtering noise and addressing distance dilution.

\begin{table}[h]
\centering
\small
\setlength{\tabcolsep}{5pt}
\begin{tabular}{l cc cc}
\toprule
\multirow{2}{*}{\textbf{Variant}} & \multicolumn{2}{c}{\textbf{ZH}} & \multicolumn{2}{c}{\textbf{EN}} \\
\cmidrule(lr){2-3} \cmidrule(lr){4-5}
& \textbf{F1} & $\Delta$ & \textbf{F1} & $\Delta$ \\
\midrule
\textbf{TCDA (Full)} & \textbf{44.35} & - & \textbf{39.69} & - \\
\quad w/o TC-DAG & 43.78 & -0.57 & 38.78 & -0.91 \\
\quad w/o D-RoPE & 43.74 & -0.61 & 38.65 & -1.04 \\
\quad w/o Both   & 43.29 & -1.06 & 38.14 & -1.55 \\
\bottomrule
\end{tabular}
\caption{Ablation results (Micro F1).}
\label{tab:ablation}
\end{table}

\subsection{Further Analysis}
\paragraph{Parameter Sensitivity. }
We investigate the impact of the TC-DAG layer $L$ and the speaker window size $w$ on the performance, as shown in Table \ref{tab:param_sensitivity}. All other hyperparameters (including the standard RoPE baseline values) are kept constant to ensure the fairness of the comparison.

\begin{table}[h]
\centering
\small
\setlength{\tabcolsep}{8pt}
\begin{tabular}{lc | lc}
\toprule
\textbf{Layers ($L$)} & \textbf{F1} & \textbf{Window ($w$)} & \textbf{F1} \\
\midrule
$L=1$ & 43.41 & $w=1$  & 42.34 \\
$L=2$ & 43.74 & $w=2$ & 43.74 \\
$L=3$ & 43.18 & $w=3$ & 43.74 \\
$L=4$ & 43.26 & $w=4$ & 43.74 \\
\bottomrule
\end{tabular}
\caption{Parameter sensitivity on ZH dataset.}
\label{tab:param_sensitivity}
\end{table}

The best performance can be achieved when $L=2$, and increasing $L$ can lead to a decrease in performance due to the over-smoothing effect. Dense connection ($w \geq 2$) is always superior to sparse connection ($w=1$), as it can facilitate direct information transmission from the root sentence to subsequent nodes, thereby maintaining the overall discourse intention in the case of long-distance attenuation. Performance saturates at $w \geq 2$ as the window size often exceeds the actual thread length.

\paragraph{Generality of D-RoPE.}
To verify the universality of D-RoPE, we replace the standard RoPE with our D-RoPE in the competitive benchmark models (i.e., MVQPN and DMIN). As shown in Table \ref{tab:generality}, D-RoPE consistently improves performance across different architectures and languages. Notably, for MVQPN, its micro F1 value increases by 1.84\% on the ZH dataset and 0.80\% on the EN dataset. This significant improvement indicates that D-RoPE effectively overcomes the limitations of the base model in capturing multi-scale positional dependencies. Moreover, the continuous improvement of DMIN further confirms that D-RoPE is a robust, model-independent plugin that can alleviate the Distance Dilution problem.

\begin{table}[h]
\centering
\small
\setlength{\tabcolsep}{4pt}
\begin{tabular}{l cc cc}
\toprule
\multirow{2}{*}{\textbf{Base Model}} & \multicolumn{2}{c}{\textbf{ZH}} & \multicolumn{2}{c}{\textbf{EN}} \\
\cmidrule(lr){2-3} \cmidrule(lr){4-5}
& \textbf{F1} & \textbf{Gain} & \textbf{F1} & \textbf{Gain} \\
\midrule
MVQPN \cite{DiaASQ} & 35.68 & - & 35.62 & - \\
\quad + D-RoPE & 37.52 & $\uparrow$ 1.84 & 36.42 & $\uparrow$ 0.80 \\
\midrule
DMIN \cite{DMIN} & 43.29 & - & 38.14 & - \\
\quad + D-RoPE & 43.78 & $\uparrow$ 0.49 & 38.78 & $\uparrow$ 0.64 \\
\bottomrule
\end{tabular}
\caption{Generality of D-RoPE on different baselines (Micro F1).}
\label{tab:generality}
\end{table}

\paragraph{Effectiveness of TC-DAG Structure.}
As shown in Table \ref{tab:dag_comparison}, to verify the necessity of our topological consistency design, we compare the proposed TC-DAG with two structural variants:
(1) Reply-GCN, which only builds an undirected graph based on reply dependencies, ignoring speaker relationships and directionality;
(2) Standard DAG, which follows the chronological order and distinguishes edges by speakers. We use the standard RoPE method in all variants.

Without thread isolation (standard DAG), unrelated thread interference degrades performance below the reply-GCN baseline on the EN dataset. Conversely, TC-DAG eliminates this interference by integrating chronological order with strict topology, achieving superior results across all metrics.

\begin{table}[h]
\centering
\small
\setlength{\tabcolsep}{10pt}
\begin{tabular}{l cc}
\toprule
\textbf{Graph Structure} & \textbf{ZH} & \textbf{EN} \\
\midrule
Reply-GCN (Undirected) & 43.29 & 38.14 \\
Standard DAG  & 43.48 & 37.57 \\
\textbf{TC-DAG (Ours)} & \textbf{43.74} & \textbf{38.65} \\
\bottomrule
\end{tabular}
\caption{Impact of graph construction strategies (Micro F1).}
\label{tab:dag_comparison}
\end{table}

\section{Conclusion}
We propose the TCDA framework to address the structural noise and scale mismatch issues in DiaASQ. We introduce TC-DAG to filter out irrelevant branches by implementing topological constraints, and introduce D-RoPE to solve the Distance Dilution by aligning the semantic granularity with the hierarchical structure of the separated subspace. TCDA achieves SOTA results in two benchmarks. The generalization ability of D-RoPE further highlights its potential as a model-agnostic plugin suitable for a wider range of multi-turn dialogue tasks.

\section*{Acknowledgments}
This work was supported by the National Natural Science Foundation of China Project (No. 62372078).
\bibliographystyle{named}
\bibliography{ijcai26}

\end{document}